\documentclass[letterpaper, 10 pt, conference]{ieeeconf}  

\IEEEoverridecommandlockouts                              

\usepackage{graphicx}                   
\usepackage{mathptmx}                   
\usepackage{times}                      
\usepackage{amsmath}                    
\usepackage{amssymb}                    
\usepackage{algorithm}
\usepackage{algorithmic}
\usepackage{subfig}
\usepackage{balance}
\graphicspath{{figures/}}

\DeclareMathOperator*{\argmax}{argmax}

\newcommand{\comment}[1]{}

\title{\LARGE \bf
Inverse Reinforcement Learning in Large State Spaces via Function Approximation
}

\author{Kun Li$^{1}$, Joel W. Burdick$^{1}$
\thanks{*This work was
supported by the National Institutes of Health, NIBIB.} \thanks{$^{1}$Kun Li and Joel W. Burdick are
with Department of Mechanical and Civil Engineering, California Institute of Technology, Pasadena,
CA 91125, USA {\tt\small kunli@caltech.edu}
}%
}
\begin{document}
\maketitle
\thispagestyle{empty}
\pagestyle{empty}

\begin{abstract}
  This paper introduces a new method for inverse reinforcement learning in large-scale and
  high-dimensional state spaces.  To avoid solving the computationally expensive reinforcement
  learning problems in reward learning, we propose a function approximation method to ensure that
  the Bellman Optimality Equation always holds, and then estimate a function to maximize the
  likelihood of the observed motion. The time complexity of the proposed method is linearly
  proportional to the cardinality of the action set, thus it can handle large state spaces
  efficiently. We test the proposed method in a simulated environment, and show that it is more
  accurate than existing methods and significantly better in scalability. We also show that the
  proposed method can extend many existing methods to high-dimensional state spaces.  We then apply
  the method to evaluating the effect of rehabilitative stimulations on patients with spinal cord
  injuries based on the observed patient motions.
\end{abstract}

\section{Introduction}
\label{irl::intro}
\begin{figure}
  \centering
  \subfloat[A patient sitting on a sensing device\label{irl::game1}]{
    \includegraphics[width=0.2\textwidth]{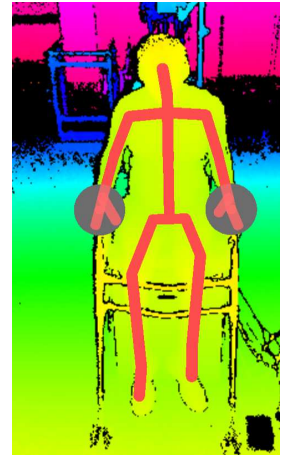}
  }
  \qquad
  \subfloat[Instructions on movement directions\label{irl::game2}]{
    \includegraphics[width=0.2\textwidth]{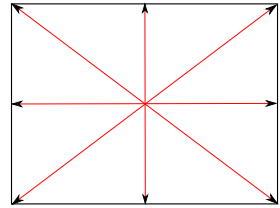}
  }
  ~
  \subfloat[The patient's COP trajectories during the movement\label{irl::game3}]{
    \includegraphics[width=0.2\textwidth]{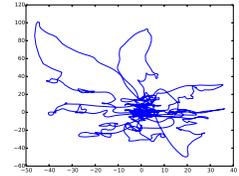}
  }
  \caption{Rehabilitative game and observed trajectories: in Figure \ref{irl::game1}, the patient
  sits on a sensing device, and then moves to follow the instructed directions in Figure
  \ref{irl::game2}. Figure  \ref{irl::game3} shows the patient's center-of-pressure (COP) during
  the movements.}
  \label{irl:game}
\end{figure}

Approximately 350,000 Americans suffer from serious spinal cord injuries (SCI), resulting in loss of
some voluntary motion control. Recently, epidural and transcutaneous spinal stimulation have proven
to be promising methods for regaining motor function. To find the optimal stimulation signal, it is
necessary to quantitatively measure the effects of different stimulations on a patient. Since motor
function is our concern, we mainly study the effects of stimulations on patient motion, represented
by a sequence of poses captured by motion sensors. One typical experiment setting is shown in Figure
\ref{irl:game}, where a patient moves to follow a physician's instructions, and a sensor records the
patient's center-of-pressure (COP) continuously. This study will assist our design of stimulating
signals, as well as advancing the understanding of patient motion with spinal cord injuries.

We assume the stimulating signals will alter the patient's initial preferences over poses,
determined by body weight distribution, spinal cord injuries, gravity, etc., and an accurate
estimation of the preference changes will reveal the effect of spinal stimulations on spinal cord
injuries, as other factors are assumed to be invariant to the stimulations. To estimate the
patient's preferences over different poses, the most straightforward approach is counting the pose
visiting frequencies from the motion, assuming that the preferred poses are more likely to be
visited.  However, the patient may visit an undesired pose to follow the instructions or to change
into a subsequent preferred poses, making preference estimation inaccurate without regarding the
context.

In this work, we formulate the patient's motion as a Markov Decision Process, where each state
represents a pose, and its reward value encodes all the immediate factors motivating the patient to
visit this state, including the pose preferences and the physician's instructions. With this
formulation, we adopt inverse reinforcement learning (IRL) algorithms to estimate the reward value
of each state from the observed motion of the patient.

Existing solutions of the IRL problem mainly work on small-scale problems, by collecting a set of
observations for reward estimation and using the estimated reward afterwards.  For example, the
methods in \cite{irl::irl1,irl::irl2, irl::subgradient} estimate the agent's policy from a set of
observations, and estimate a reward function that leads to the policy.  The method in
\cite{irl::maxentropy} collects a set of trajectories of the agent, and estimates a reward function
that maximizes the likelihood of the trajectories. However, the state space of human motion is huge
for non-trivial analysis, and these methods cannot handle it well due to the reinforcement learning
problem in each iteration of reward estimation. Several methods \cite{irl::guidedirl,irl::relative}
solve the problem by approximating the reinforcement learning step, at the expense of a
theoretically sub-optimal solution.

The problem can be simplified under the condition that the transition model and the action set
remain unchanged for the subject, thus each reward function leads to a unique optimal value
function. Based on this assumption, we propose a function approximation method that learns the
reward function and the optimal value function, but without the computationally expensive
reinforcement learning steps, thus it can be scaled to a large state space. We find that this
framework can also extend many existing methods to high-dimensional state spaces.

The paper is organized as follows. We review existing work on inverse reinforcement learning in
Section \ref{irl::related}, and formulate the function approximation inverse reinforcement learning
method in large state spaces in \ref{irl::largeirl}. A simulated experiment and a clinical
experiment are shown in Section \ref{irl::experiments}, with conclusions in Section
\ref{irl::conclusions}.

\section{Related Works}
\label{irl::related}
The idea of inverse optimal control is proposed by Kalman \cite{irl::kalman}, white the inverse
reinforcement learning problem is firstly formulated in \cite{irl::irl1}, where the agent observes
the states resulting from an assumingly optimal policy, and tries to learn a reward function that
makes the policy better than all alternatives. Since the goal can be achieved by multiple reward
functions, this paper tries to find one that maximizes the difference between the observed policy
and the second best policy. This idea is extended by \cite{irl::maxmargin}, in the name of
max-margin learning for inverse optimal control. Another extension is proposed in \cite{irl::irl2},
where the purpose is not to recover the real reward function, but to find a reward function that
leads to a policy equivalent to the observed one, measured by the amount of rewards collected by
following that policy.

Since a motion policy may be difficult to estimate from observations, a behavior-based method is
proposed in \cite{irl::maxentropy}, which models the distribution of behaviors as a maximum-entropy
model on the amount of reward collected from each behavior. This model has many applications and
extensions. For example, \cite{irl::sequence} considers a sequence of changing reward functions
instead of a single reward function. \cite{irl::gaussianirl} and \cite{irl::guidedirl} consider
complex reward functions, instead of linear one, and use Gaussian process and neural networks,
respectively, to model the reward function. \cite{irl::pomdp} considers complex environments,
instead of a well-observed Markov Decision Process, and combines partially observed Markov Decision
Process with reward learning. \cite{irl::localirl} models the behaviors based on the local
optimality of a behavior, instead of the summation of rewards.  \cite{irl::deepirl} uses a
multi-layer neural network to represent nonlinear reward functions.

Another method is proposed in \cite{irl::bayirl}, which models the probability of a behavior as the
product of each state-action's probability, and learns the reward function via maximum a posteriori
estimation. However, due to the complex relation between the reward function and the behavior
distribution, the author uses computationally expensive Monte-Carlo methods to sample the
distribution. This work is extended by \cite{irl::subgradient}, which uses sub-gradient methods to
simplify the problem.  Another  extensions is shown in \cite{irl::bayioc}, which tries to find a
reward function that matches the observed behavior. For motions involving multiple tasks and varying
reward functions, methods are developed in \cite{irl::multirl1} and \cite{irl::multirl2}, which try
to learn multiple reward functions. 

Most of these methods need to solve a reinforcement learning problem in each step of reward
learning, thus practical large-scale application is computationally infeasible. Several methods are
applicable to large-scale applications. The method in \cite{irl::irl1} uses a linear approximation
of the value function, but it requires a set of manually defined basis functions. The methods in
\cite{irl::guidedirl,irl::relative} update the reward function parameter by minimizing the relative
entropy between the observed trajectories and a set of sampled trajectories based on the reward
function, but they require a set of manually segmented trajectories of human motion, where the
choice of trajectory length will affect the result. Besides, these methods solve large-scale
problems by approximating the Bellman Optimality Equation, thus the learned reward function and Q
function are only approximately optimal. We propose an approximation method that guarantees the
optimality of the learned functions as well as the scalability to large state space problems.

\section{Function Approximation Inverse Reinforcement Learning}
\label{irl::largeirl}
\subsection{Markov Decision Process}
A Markov Decision Process is described with the following variables:
\begin{itemize}
  \item $S=\{s\}$, a set of states
  \item $A=\{a\}$, a set of actions
  \item $P_{ss'}^a$, a state transition function that defines the probability that state $s$ becomes
    $s'$ after action $a$.
  \item $R=\{r(s)\}$, a reward function that defines the immediate reward of state $s$.
  \item $\gamma$, a discount factor that ensures the convergence of the MDP over an infinite
    horizon.
\end{itemize}

A motion can be represented as a sequence of state-action pairs:
\[\zeta=\{(s_i,a_i)|i=0,\cdots,N_\zeta\},\]
where $N_\zeta$ denotes the length of the motion, varying in different observations. Given the
observed sequence, inverse reinforcement learning algorithms try to recover a reward function that
explains the motion.

One key problem is how to model the action in each state, or the policy, $\pi(s)\in A$, a mapping
from states to actions. This problem can be handled by reinforcement learning algorithms, by
introducing the value function $V(s)$ and the Q-function $Q(s,a)$, described by the Bellman Equation
\cite{irl::rl}:
\begin{align}
  &V^\pi(s)=\sum_{s'|s,\pi(s)}P_{ss'}^{\pi(s)}[r(s')+\gamma*V^\pi(s')],\\
  &Q^\pi(s,a)=\sum_{s'|s,a}P_{ss'}^a[r(s')+\gamma*V^\pi(s')],
\end{align}
where $V^\pi$ and $Q^\pi$ define the value function and the Q-function under a policy $\pi$.

For an optimal policy $\pi^*$, the value function and the Q-function should be maximized on every
state. This is described by the Bellman Optimality Equation \cite{irl::rl}:
\begin{align}
  &V^*(s)=\max_{a\in A}\sum_{s'|s,a}P_{ss'}^a[r(s')+\gamma*V^*(s')],\\
  &Q^*(s,a)=\sum_{s'|s,a}P_{ss'}^a[r(s')+\gamma*\max_{a'\in A}Q^*(s',a')].
\end{align}

In typical inverse reinforcement learning algorithms, the Bellman Optimality Equation needs to be
solved once for each parameter updating of the reward function, thus it is computationally
infeasible when the state space is large. While several existing approaches solve the problem at the
expense of the optimality, we propose an approximation method to avoid the problem.

\subsection{Function Approximation Framework}
Given the set of actions and the transition probability, a reward function leads to a unique optimal
value function. To learn the reward function from the observed motion, instead of directly
learning the reward function, we use a parameterized function, named as \textit{VR function}, to
represent the summation of the reward function and the discounted optimal value function:
\begin{equation}
  f(s,\theta)=r(s)+\gamma*V^*(s),
  \label{equation:approxrewardvalue}
\end{equation}
where $\theta$ denotes the parameter of \textit{VR function}. The function value of a state is named
as \textit{VR value}.

Substituting Equation \eqref{equation:approxrewardvalue} into Bellman Optimality Equation, the
optimal Q function is given as:
\begin{equation}
  Q^*(s,a)=\sum_{s'|s,a}P_{ss'}^af(s',\theta),
  \label{equation:approxQ}
\end{equation}
the optimal value function is given as:
\begin{align}
  V^*(s)&=\max_{a\in A}Q^*(s,a)\nonumber\\
        &=\max_{a\in A}\sum_{s'|s,a}P_{ss'}^af(s',\theta),
  \label{equation:approxV}
\end{align}
and the reward function can be computed as:
\begin{align}
  r(s)&=f(s,\theta)-\gamma*V^*(s)\nonumber\\
      &=f(s,\theta)-\gamma*\max_{a\in A}\sum_{s'|s,a}P_{ss'}^af(s',\theta).
  \label{equation:approxR}
\end{align}

This approximation method is related to value function approximation method in reinforcement
learning, but the proposed method can compute the reward function without solving a set of linear
equations in stochastic environments. 

Note that this formulation can be generalized to other extensions of Bellman Optimality Equation by
replacing the $max$ operator with other types of Bellman backup operators. For example,
$V^*(s)=\log_{a\in A}\exp Q^*(s,a)$ is used in the maximum-entropy method\cite{irl::maxentropy};
$V^*(s)=\frac{1}{k}\log_{a\in A}\exp k*Q^*(s,a)$ is used in Bellman Gradient Iteration
\cite{irl::BGI}.  

For any \textit{VR function} $f$ and any parameter $\theta$, the optimal Q function $Q^*(s,a)$,
optimal value function $V^*(s)$, and reward function $r(s)$ constructed with Equation
\eqref{equation:approxQ}, \eqref{equation:approxV}, and \eqref{equation:approxR} always meet the
Bellman Optimality Equation. Under this condition, we try to recover a parameterized function
$f(s,\theta)$ that best explains the observed motion $\zeta$ based on a predefined motion model.

Combined with different Bellman backup operators, this formulation can extend many existing methods
to high-dimensional spaces, like the motion model based on the value function in
\cite{irl::motionvalue}, $p(a|s)=-v(s)-\log\sum_k p_{s,k}\exp(-v(k))$, the reward function in
\cite{irl::maxentropy}, $p(a|s)=\exp{Q(s,a)-V(s)}$, and the Q function in \cite{irl::bayirl}. The
main limitation is the assumption of a known transition model $P_{ss'}^a$, but it only requires a
partial model on the experienced states rather than a full environment model, and it can be learned
independently in an unsupervised way.

To demonstrate the usage of the framework, this work chooses $max$ as the Bellman backup operator
and a motion model $p(a|s)$ based on the optimal Q function $Q^*(s,a)$ \cite{irl::bayirl}:
\begin{equation}
  P(a|s)=\frac{\exp{b*Q^*(s,a)}}{\sum_{\tilde{a}\in
  A}\exp{b*Q^*(s,\tilde{a})}},
  \label{equation:motionmodel}
\end{equation} 
where $b$ is a parameter controlling the degree of confidence in the agent's ability to choose
actions based on Q values. In the remaining sections, we use $Q(s,a)$ to denote the optimal Q values
for simplified notations.

\subsection{Function Approximation with Neural Network}
Assuming the approximation function is a neural network, the parameter $\theta=\{w,b\}$-weights and
biases-in Equation \eqref{equation:approxrewardvalue} can be estimated from the observed sequence
of state-action pairs $\zeta$ via maximum-likelihood estimation:
\begin{equation}
  \label{equation:maxtheta}
  \theta=\argmax_{\theta}\log{P(\zeta|\theta)},
\end{equation}
where the log-likelihood of $P(\zeta|\theta)$ is given by:
\begin{align}
  L_{nn}(\theta)&=\log{P(\zeta|\theta)}\nonumber\\
           &=\log{\prod_{(s,a)\in \zeta} P(a|\theta;s)}\nonumber\\
           &=\log{\prod_{(s,a)\in \zeta} \frac{\exp{b*Q^*(s,a)}}{\sum_{\hat{a}\in A}\exp{b*Q^*(s,\hat{a})}}
}\nonumber\\
  &=\sum_{(s,a)\in\zeta}(b*Q(s,a)-\log{\sum_{\hat{a}\in
  A}\exp{b*Q(s,\hat{a}))}},
  \label{equation:loglikelihood}
\end{align}
and the gradient of the log-likelihood is given by:
\begin{align}
  \nabla_\theta L_{nn}(\theta)&=\sum_{(s,a)\in\zeta}(b*\nabla_\theta Q(s,a)\nonumber\\
  &-b*\sum_{\hat{a}\in
  A}P((s,\hat{a})|r(\theta))\nabla_\theta Q(s,\hat{a})).
  \label{equation:loglikelihoodgradient}
\end{align}

With a differentiable approximation function, 
\[\nabla_\theta Q(s,a)=\sum_{s'|s,a}P_{ss'}^a\nabla_\theta f(s',\theta),\]
and 
\begin{align}
  \nabla_\theta L_{nn}(\theta)&=\sum_{(s,a)\in\zeta}(b*\sum_{s'|s,a}P_{ss'}^a\nabla_\theta f(s',\theta)\nonumber\\
  &-b*\sum_{\hat{a}\in
  A}P((s,\hat{a})|r(\theta))\sum_{s'|s,a}P_{ss'}^a\nabla_\theta f(s',\theta)),
  \label{equation:loglikelihoodgradient}
\end{align}
where $\nabla_\theta f(s',\theta)$ denotes the gradient of the neural network output with respect to neural
network parameter $\theta=\{w,b\}$.

If the \textit{VR function} $f(s,\theta)$ is linear, the objective function in Equation
\eqref{equation:loglikelihood} is concave, and a global optimum exists. However, a multi-layer
neural network works better to handle the non-linearity in approximation and the high-dimensional
state space data.

A gradient ascent method can be used to learn the parameter $\theta$:
\begin{equation}
  \label{equation:gradientascent}
  \theta=\theta+\alpha*\nabla_\theta L_{nn}(\theta),
\end{equation}
where $\alpha$ is the learning rate.

When the method converges, we can compute the optimal Q function, the optimal value function, and the
reward function based on Equation \eqref{equation:approxrewardvalue}, \eqref{equation:approxQ},
\eqref{equation:approxV}, and \eqref{equation:approxR}. The algorithm under a neural network-based
approximation function is shown in Algorithm \ref{alg:nnapprox}.

This method does not involve solving the MDP problem for each updated parameter $\theta$, and
large-scale state spaces can be easily handled by an approximation function based on a deep neural
network.
\begin{algorithm}[tb]
  \caption{Function Approximation IRL with Neural Network}
  \label{alg:nnapprox}
\begin{algorithmic}[1]
  \STATE Data: {$\zeta,S,A,P,\gamma,b,\alpha$}
  \STATE Result: {optimal value $V[S]$, optimal action value $Q[S,A]$, reward value $R[S]$}
  \STATE create variable $\theta=\{W,b\}$ for a neural network
  \STATE build $f[S,\theta]$ as the output of the neural network
  \STATE build $Q[S,A]$, $V[S]$, and $R[S]$ based on Equation \eqref{equation:approxrewardvalue},
  \eqref{equation:approxQ}, \eqref{equation:approxV}, and \eqref{equation:approxR}.
  \STATE build loglikelihood $L_{nn}[\theta]$ based on $\zeta$ and $Q[S,A]$
  \STATE compute gradient $\nabla_\theta L_{nn}[\theta]$
  \STATE initialize $\theta$
  \WHILE{not converging}
    \STATE $\theta=\theta+\alpha*\nabla_\theta L_{nn}[\theta]$
  \ENDWHILE
  \STATE evaluate {optimal value $V[S]$, optimal action value $Q[S,A]$, reward value $R[S]$}
  \STATE return $R[S]$
\end{algorithmic}
\end{algorithm}

\subsection{Function Approximation with Gaussian Process}
Assuming the \textit{VR function} $f$ is a Gaussian process (GP) parameterized by $\theta$, the
posterior distribution is similar to the distribution in \cite{irl::gaussianirl}:
\begin{align}
  P(\theta,f_u|S_u,\zeta)
  &\propto P(\zeta,f_u,\theta|S_u)\\\nonumber
  &= \int_{f_S}\underbrace{P(\zeta|f_S)}_\text{IRL}\underbrace{P(f_S|f_u,\theta,S_u)}_\text{GP
  posterior}df_S\underbrace{P(f_u,\theta|S_u)}_\text{GP prior},
\end{align}
where $S_u$ denotes a set of supporting states for sparse Gaussian approximation
\cite{irl::sparsegaussian}, $f_u$ denotes the \textit{VR values} of $S_u$, $f_S$ denotes the
\textit{VR values} of the whole set of states, and $\theta$ denotes the parameter of the Gaussian
process.

Without a closed-form integration, we use the mean function of the Gaussian posterior as the
\textit{VR value}:
\begin{align}
  P(\zeta,f_u,\theta|S_u)=P(\zeta|\bar{f_S})P(f_u,\theta|S_u),
  \label{equation:gpfairl}
\end{align}
where $\bar{f_S}$ denotes the mean function.

Given a kernel function $k(x_i,x_j,\theta)$, the log-likelihood function is given as:
\begin{align}
  &L_{gp}(\theta,f_u)\label{eq:gausslikelihood}\\&=\log P(\zeta|\bar{f_S})+\log P(f_u,\theta|S_u)\\
               &=b*\sum_{(s,a)\in\zeta}(\sum_{s'|s,a}P_{ss'}^a\bar{f(s')}
  -\log{\sum_{\hat{a}\in A}\exp{\sum_{s'|s,\hat{a}}P_{ss'}^{\hat{a}}\bar{f(s'))}}}\label{eq:irlterm}\\
  &-\frac{f_u^TK_{S_u,S_u}^{-1}f_u}{2}-\frac{\log|K_{S_u,S_u}|}{2}-\frac{n\log
  2\pi}{2}\label{eq:gaussterm}
\\  &+\log P(\theta),\label{eq:priorterm}
\end{align}
where $K$ denotes the covariance matrix computed with the kernel function,
$\bar{f(s)}=K_{s,S_u}^TK_{S_u,S_u}^{-1}f_u$ denotes the \textit{VR value} with the mean function
$\bar{f_S}$, expression \eqref{eq:irlterm} is the IRL likelihood, expression \eqref{eq:gaussterm} is
the Gaussian prior likelihood, and expression \eqref{eq:priorterm} is the kernel parameter prior.

The parameters $\theta,f_u$ can be similarly learned with gradient methods. It has similar
properties with neural net-based approach, and the full algorithm is shown in Algorithm
\ref{alg:gpapprox}.
\begin{algorithm}[tb]
  \caption{Function Approximation IRL with Gaussian Process}
  \label{alg:gpapprox}
\begin{algorithmic}[1]
  \STATE Data: {$\zeta,S,A,P,\gamma,b,\alpha$}
  \STATE Result: {optimal value $V[S]$, optimal action value $Q[S,A]$, reward value $R[S]$}
  \STATE create variable $\theta$ for a kernel function and $f_u$ for supporting points
  \STATE compute $\bar{f(s,\theta,f_u)}=K_{s,S_u}^TK_{S_u,S_u}^{-1}f_u$
  \STATE build $Q[S,A]$, $V[S]$, and $R[S]$ based on Equation \eqref{equation:approxrewardvalue},
  \eqref{equation:approxQ}, \eqref{equation:approxV}, and \eqref{equation:approxR}.
  \STATE build loglikelihood $L_{gp}[\theta,f_u]$ based on Equation \eqref{eq:gausslikelihood}.
  \STATE compute gradient $\nabla_{\theta,f_u} L_{gp}[\theta,f_u]$
  \STATE initialize $\theta,f_u$
  \WHILE{not converging}
    \STATE $[\theta,f_u]=[\theta,f_u]+\alpha*\nabla_{\theta,f_u} L_{gp}[\theta,f_u]$
  \ENDWHILE
  \STATE evaluate {optimal value $V[S]$, optimal action value $Q[S,A]$, reward value $R[S]$}
  \STATE return $R[S]$
\end{algorithmic}
\end{algorithm}

\section{Experiments}
\label{irl::experiments}
We use a simulated environment to compare the proposed methods with existing methods and demonstrate
the accuracy and scalability of the proposed solution, then we show how the function approximation
framework can extend existing methods to large state spaces.  In the end, we apply the proposed
method to a clinical application.
\subsection{Simulated Environment}
\begin{figure}
  \centering
 \includegraphics[width=0.2\textwidth]{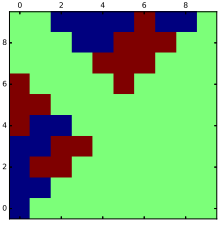}
  \caption{An example of a reward table for one objectworld mdp on a $10\times 10$ grid: it depends
  on randomly placed objects.}
  \label{fig:objectworld}
\end{figure}

The simulated environment is an objectworld mdp \cite{irl::gaussianirl}. It is a $N*N$ grid, but
with a set of objects randomly placed on the grid. Each object has an inner color and an outer
color, selected from a set of possible colors, $C$. The reward of a state is positive if it is
within 3 cells of outer color $C1$ and 2 cells of outer color $C2$, negative if it is within 3 cells
of outer color $C1$, and zero otherwise. Other colors are irrelevant to the ground truth reward. One
example of the reward values is shown in Figure \ref{fig:objectworld}. In this work, we place two
random objects on a $5*5$ grid, and the feature of a state describes its discrete distance to each
inter color and outer color in $C$.

We evaluate the proposed method in three aspects. First, we compare its accuracy in reward learning
with other methods. We generate different sets of trajectory samples, and implement the
maximum-entropy method in \cite{irl::maxentropy}, deep inverse reinforcement learning method in
\cite{irl::deepirl}, and Bellman Gradient Iteration approaches \cite{irl::BGI}. The \textit{VR
function} based on a neural network has five-layers, where the number of nodes in the first four
layers equals to the feature dimensions, and the last layer outputs a single value as the summation
of the reward and the optimal value. The \textit{VR function} based on a Gaussian process uses an
automatic relevance detection (ARD) kernel \cite{irl::gaussianml} and an uninformed prior, and the
supporting points are randomly picked. The accuracy is defined as the correlation coefficient
between the ground truth reward value and the learned reward value.

\begin{figure}
  \centering
  \includegraphics[width=0.4\textwidth]{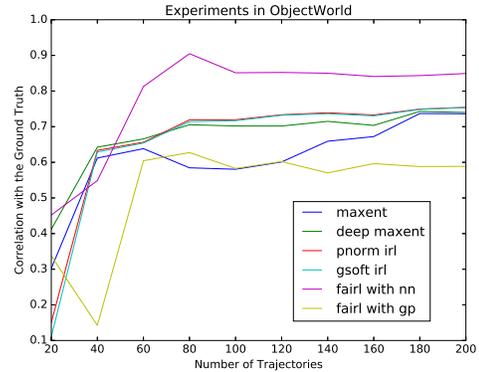}
  \caption{Accuracy comparison with different numbers of observations: "maxent" denotes maximum
  entropy method; "deep maxent" denotes the deep inverse reinforcement learning approach, "pnorm
  irl" and "gsoft irl" denote Bellman Gradient Iteration method; "fairl with nn" denotes the
  function approximation inverse reinforcement learning with a neural network; "fairl with gp"
  denotes the function approximation inverse reinforcement learning with a Gaussian process.}
  \label{fig:objectaccuracy}
\end{figure}

The result is shown in Figure \ref{fig:objectaccuracy}. The accuracy is not monotonously increasing
as the number of sample grows. The reason is that a function approximator based on a large neural
network will overfit the observed trajectory, which may not reflect the true reward function
perfectly. During reward learning, we observe that as the loglikelihood increases, the accuracy of
the updated reward function reaches the maximum after a certain number of iterations, and then
decreases to a stable value. A possible solution to this problem is early-stopping during reward
learning. For a function approximator with Gaussian process, the supporting set is important,
although an universal solution is unavailable.

Second, we evaluate the scalability of the proposed method. Since all these methods involve gradient
method, we choose different numbers of states, ranging from 25 to 9025, and compute the time for
one iteration of gradient ascent under each state size with each method. "Maxent" and "BGI" are
implemented with a mix of Python and C programming language; "DeepMaxent" is implemented with
Theano, and "FAIRL" is implemented with Tensorflow. They all have C programming language in the
backend and Python in the forend. 

\begin{table}
  \caption{The computation time (second) of one iteration of gradient method under different number
  of states with different methods: "Maxent" denotes maximum entropy method, "DeepMaxent" denotes
  the deep inverse reinforcement learning approach, "BGI" denotes Bellman Gradient Iteration method,
  and "FAIRLNN" and "FAIRLGP" denote the function approximation inverse reinforcement learning.}
  \label{tab:time}
  \begin{tabular}{|c|c|c|c|c|c|}
    \hline
    States (\#)&Maxent&DeepMaxent&BGI&FAIRLNN&FAIRLGP\\
    \hline
      25&0.017&0.012&0.0313&0.197&0.331\\
    \hline
      225& 1.831&0.178& 2.031& 0.397&0.721\\
    \hline
      625& 24.151&0.95& 20.963& 0.724&1.317\\
     \hline
      1225& 133.839&3.158& 102.460& 0.921&2.163\\
     \hline
      2025& 474.907&8.119& 352.007& 0.776&2.332\\
      \hline
      3025& 1319.365&20.253& 1061.147& 0.762&3.723\\
      \hline
      4225& 3030.723&59.279& 2630.309& 2.468&4.459\\
      \hline
      5625& 6197.718&101.434& 5228.343& 2.831&6.495\\
      \hline
      7225& 12234.417&229.752& 10147.628& 2.217&9.316\\
      \hline
      9025& 20941.9&10466.784& 16345.874& 3.347&12.372\\
    \hline
  \end{tabular}
\end{table}

The result is shown in Table \ref{tab:time}. Even though the computation time may be affected by
different implementations, it still shows that the  proposed method is significantly better than
the alternatives in scalability, and in practice, it can be further improved by paralleling the
computation of the reward function, the value function, and the Q function from the function
approximator. Besides, the Gaussian process-based method requires more time than the neural net,
because of the matrix inverse operations.

\begin{figure}
  \centering
  \includegraphics[width=0.4\textwidth]{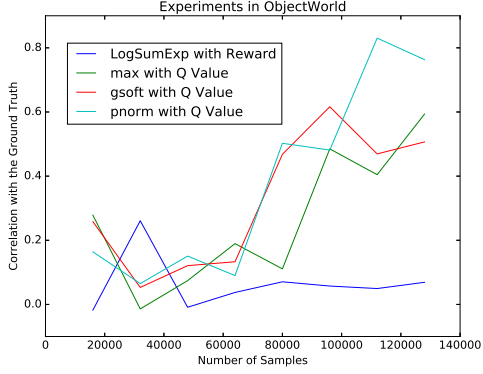}
 \caption{Reward learning accuracy of existing methods in large state spaces: "LogSumExp", "Max",
  "PNorm", and "GSoft" are the Bellman backup operators; "Reward" and "QValues" are the types of
  motion models; different combinations of extended methods are plotted. The accuracy is measured as
  the correlation between the ground truth and the recovered reward.}
  \label{fig:extendaccuracy}
\end{figure}

\begin{figure}
  \centering
  \includegraphics[width=0.4\textwidth]{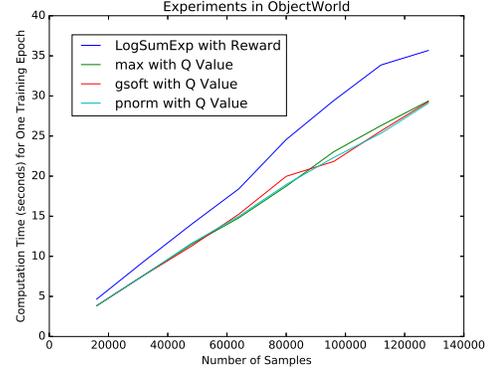}
 \caption{Computation time for one training epoch of existing methods in large state spaces:
  "LogSumExp", "Max", "PNorm", and "GSoft" are the Bellman backup operators; "Reward" and "QValues"
  are the types of motion models; different combinations of extended methods are plotted.}
  \label{fig:extendtime}
\end{figure}

Third, we demonstrate how the proposed framework extends existing methods to large-scale state
spaces.  We increase the objectworld to a $80*80$ grid, with 10 objects in 5 colors, and generate
a large sample set with size ranging from 16000 to 128000 at an interval of 16000. Then we show the
accuracy and computation time of inverse reinforcement learning with different combinations of
Bellman backup operators and motion models. The combinations include LogSumExp as Bellman backup
operator with a motion model based on the reward value \cite{irl::maxentropy} and three Bellman
backup operators ($max$, $pnorm$, $gsoft$) with a motion model based on the Q values. We do not use
even larger state spaces because the generation of trajectories from the ground truth reward function
requires a computation-intensive and memory-intensive reinforcement learning step in larger state
spaces. A three-layer neural network is adopted for function approximation, implemented with
Tensorflow on NVIDIA GTX 1080. The training is done with batch sizes 400, learning rate 0.001, and
20 training epochs are ran. The accuracy is shown in Figure \ref{fig:extendaccuracy}.  The
computation time for one training epoch is shown in Figure \ref{fig:extendtime}.

The results show that the proposed method achieves accuracy and efficiency simultaneously. In
practice, multi-start strategy may be adopted to avoid local optimum. 
\subsection{Clinical Experiment}
In the clinic, a patient with spinal cord injuries sits on a box, with a force sensor, capturing the
center-of-pressure (COP) of the patient during movement. Each experiment is composed of two
sessions, one without transcutaneous stimulation and one with stimulation. The electrodes
configuration and stimulation signal pattern are manually selected by the clinician.

In each session, the physician gives eight (or four) directions for the patient to follow, including
left, forward left, forward, forward right, right, right backward, backward, backward left, and the
patient moves continuously to follow the instruction. The physician observes the patient's behaviors
and decides the moment to change the instruction.

\begin{figure}
  \centering
  \includegraphics[width=0.4\textwidth]{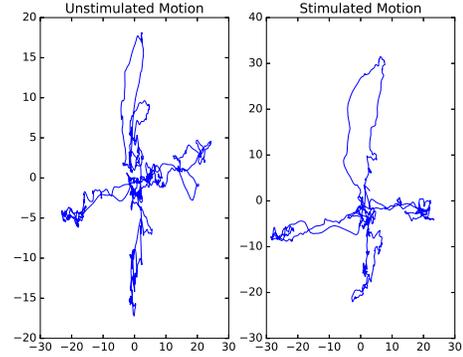}
  \caption{Patient 1 under four directional instructions: "unstimulated motion" means that the
  patient moves without transcutaneous stimulations, while "stimulated motion" represents the motion
  under stimulations.}
  \label{fig:patient1}
\end{figure}

\begin{figure}
  \centering
  \includegraphics[width=0.4\textwidth]{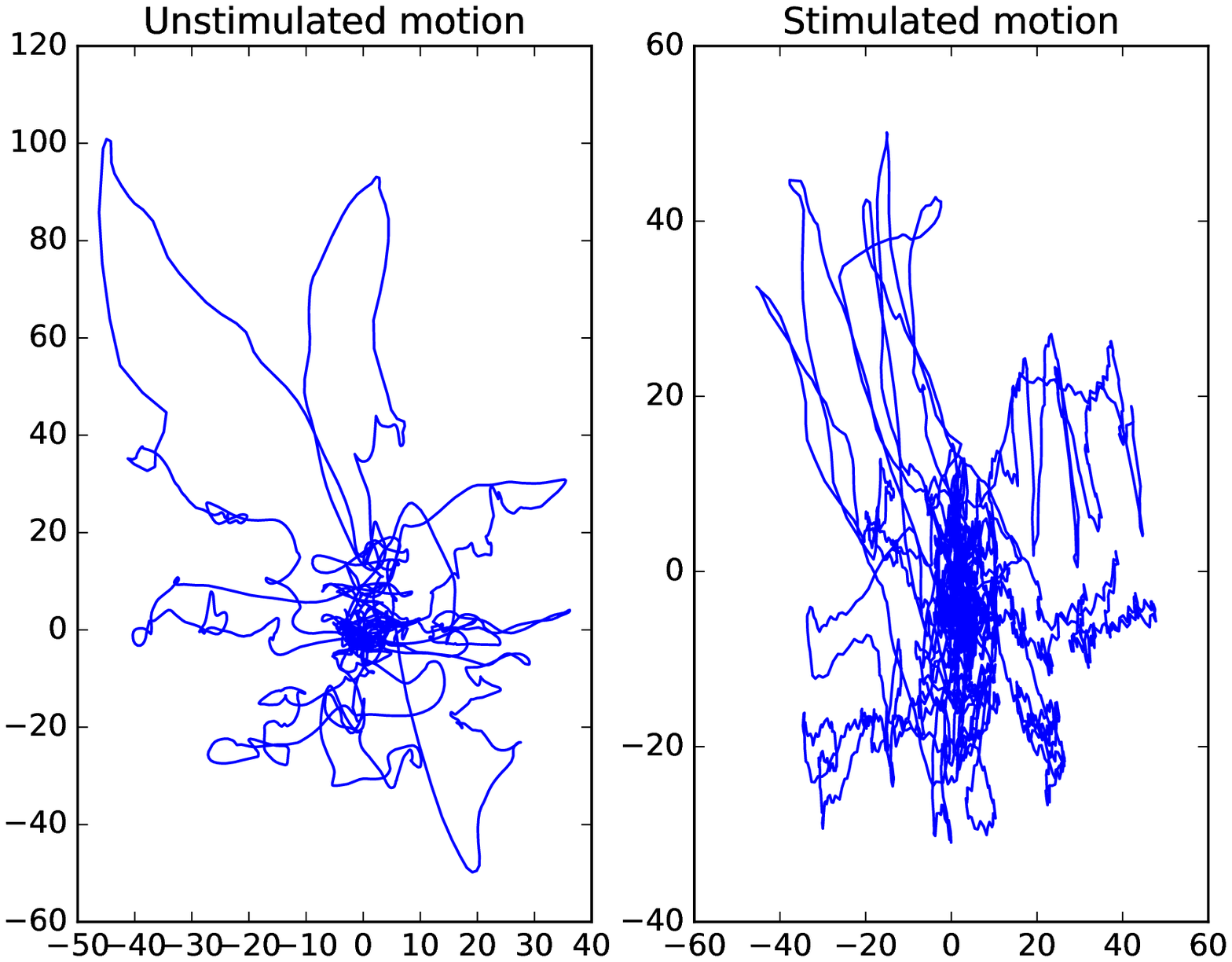}
  \caption{Patient 2 under eight directional instructions: "unstimulated motion" means that the
  patient moves without transcutaneous stimulations, while "stimulated motion" represents the motion
  under stimulations.}
  \label{fig:patient2}
\end{figure}

\begin{figure}
  \centering
  \includegraphics[width=0.4\textwidth]{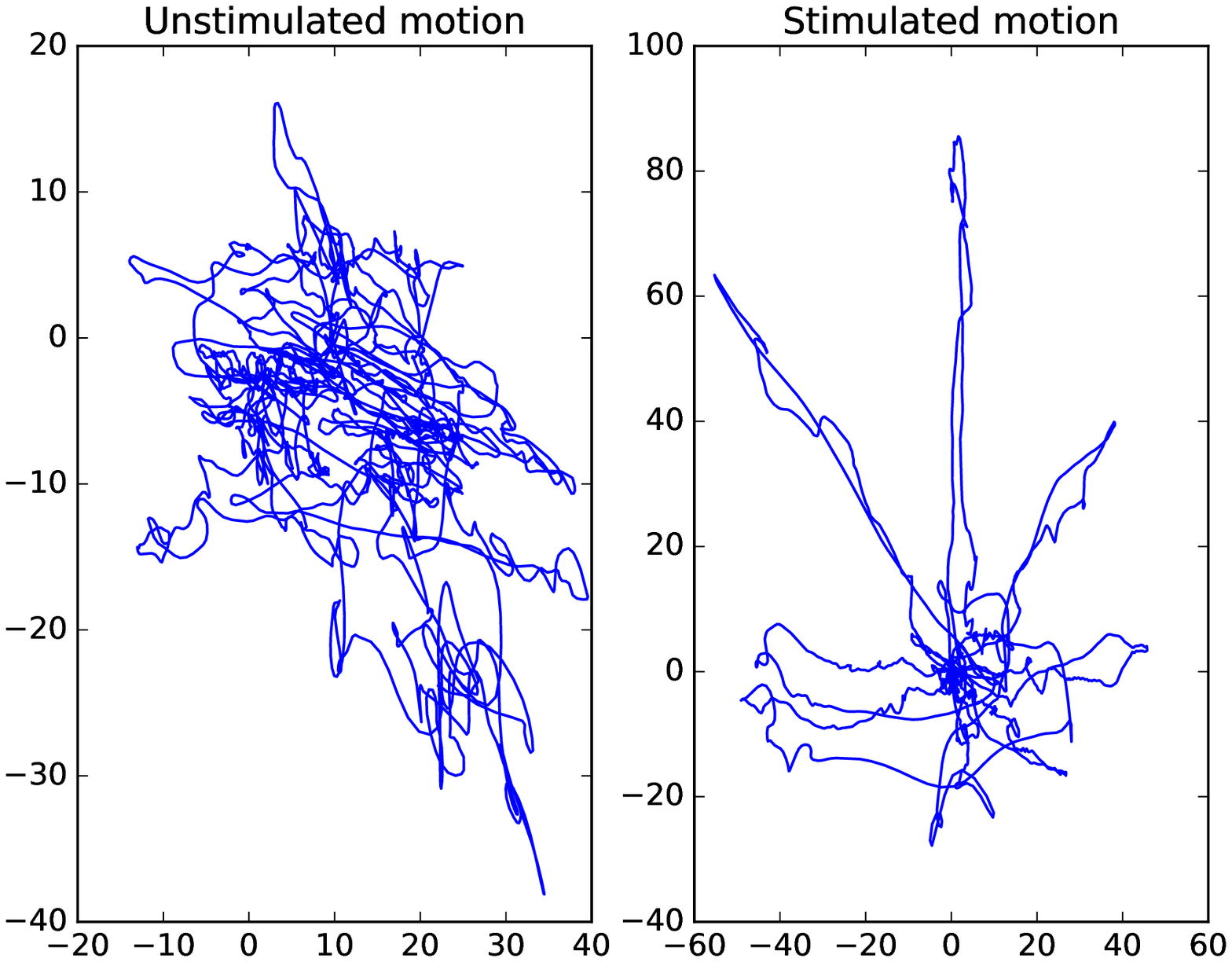}
  \caption{Patient 3 under eight directional instructions: "unstimulated motion" means that the
  patient moves without transcutaneous stimulations, while "stimulated motion" represents the motion
  under stimulations.}
  \label{fig:patient3}
\end{figure}

\begin{figure}
  \centering
  \includegraphics[width=0.4\textwidth]{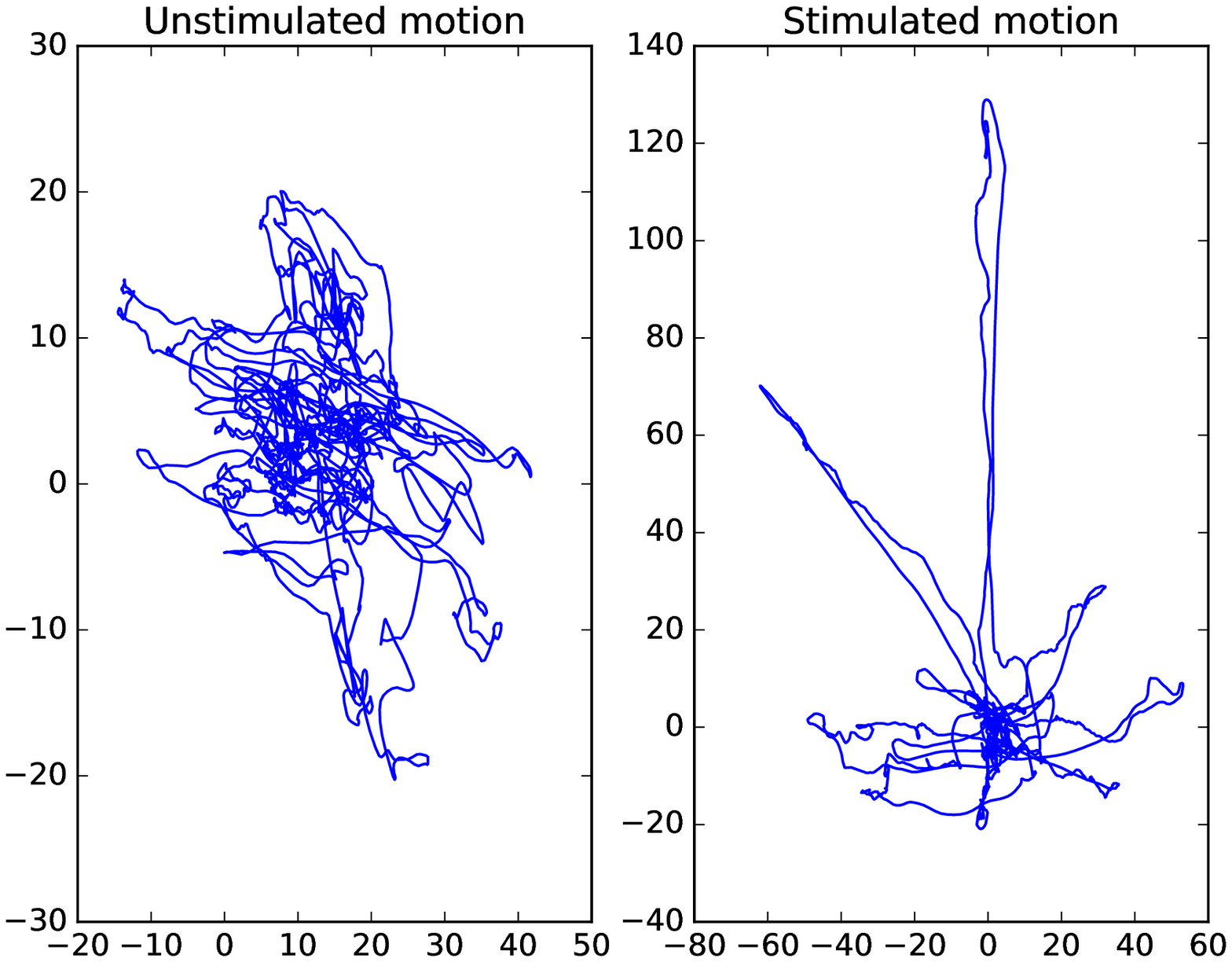}
  \caption{Patient 4 under eight directional instructions: "unstimulated motion" means that the
  patient moves without transcutaneous stimulations, while "stimulated motion" represents the motion
  under stimulations.}
  \label{fig:patient4}
\end{figure}

\begin{figure}
  \centering
  \includegraphics[width=0.4\textwidth]{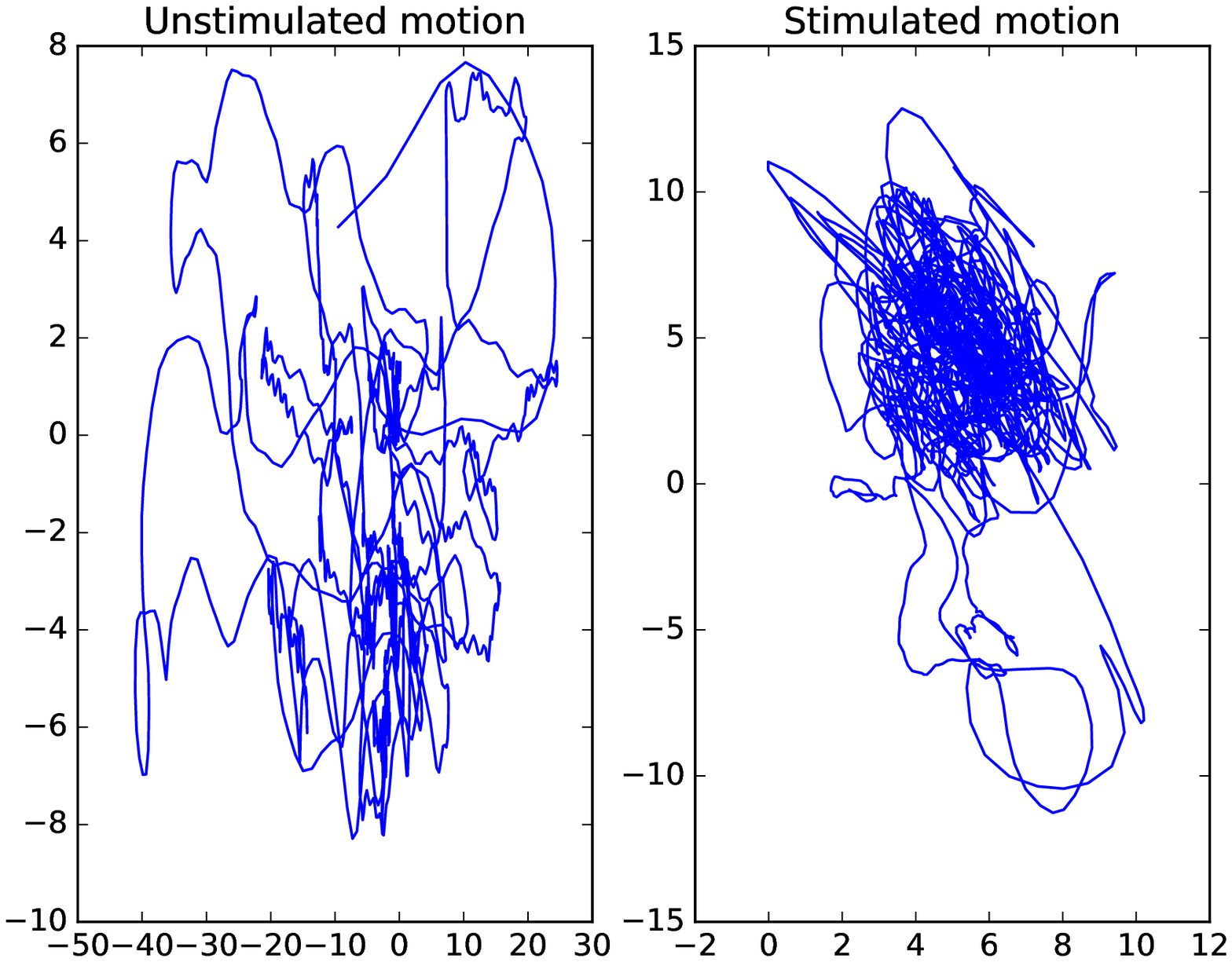}
  \caption{Patient 5 under eight directional instructions: "unstimulated motion" means that the
  patient moves without transcutaneous stimulations, while "stimulated motion" represents the motion
  under stimulations.}
  \label{fig:patient5}
\end{figure}

\begin{figure}
  \centering
  \includegraphics[width=0.4\textwidth]{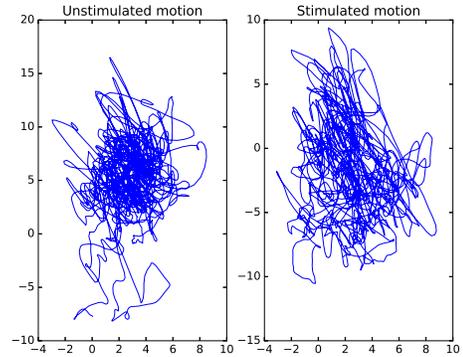}
  \caption{Patient 6 under eight directional instructions: "unstimulated motion" means that the
  patient moves without transcutaneous stimulations, while "stimulated motion" represents the motion
  under stimulations.}
  \label{fig:patient6}
\end{figure}

Six experiments are done, each with two sessions. The COP trajectories in Figure \ref{fig:patient1}
denote the case with four directional instructions; Figure \ref{fig:patient2}, \ref{fig:patient3},
\ref{fig:patient4}, \ref{fig:patient5}, and \ref{fig:patient6} denote the sessions with eight
directional instructions.


The COP sensory data from each session is discretized on a $100\times 100$ grid, which is fine
enough to capture the patient's small movements. The problem is formulated into a MDP, where each
state captures the patient's discretized location and velocity, and the set of actions changes the
velocity into eight possible directions. The velocity is represented with a two-dimensional vector
showing eight possible velocity directions. Thus the problem has 80000 states and 8 actions, and the
transition model assumes that each action will lead to one state with probability one. 
\begin{table*}
  \centering
  \caption{Evaluation of the learned rewards: "forward" etc. denote the instructed direction; "1u"
  denote the patient id "1", with "u" denoting unstimulated session and "s" denoting stimulated
  sessions. The table shows the correlation coefficient between the ideal reward and the recovered
  reward.}
  \label{tab:feature1}
 \begin{tabular}{|c|c|c|c|c|c|c|c|c|c|}
      \hline
     &forward&backward&left&right&top left&top right&bottom left&bottom right&origin\\\hline
1u&-0.352172&-0.981877&-0.511908&-0.399777&&&&&-0.0365778\\\hline
1s&-0.36437&-0.999993&-0.14757&-0.321745&&&&&0.154132\\\hline
2u&-0.459214&-0.154868&0.134229&0.181629&0.123853&0.677538&-0.398259&0.264739&-0.206476\\\hline
2s&-0.115516&-0.127179&0.569024&0.164638&0.360013&0.341521&0.0817681&0.134049&-0.00986036\\\hline
3u&0.533031&0.0364088&0.128325&-0.729293&0.397182&0.155565&-0.48818&-0.293617&-0.176923\\\hline
3s&-0.340902&-0.091139&0.344993&0.0557266&0.162783&0.740827&-0.0897398&-0.00674047&-0.414462\\\hline
4u&0.099563&-0.0965766&0.145509&-0.912844&0.250434&-0.299531&0.577489&0.134106&-0.151334\\\hline
4s&-0.258762&-0.019275&-0.263354&0.549305&0.0910128&0.755755&-0.225137&0.289126&-0.216737\\\hline
5u&0.287442&0.0859648&-0.368503&0.504589&-0.297166&0.401829&0.0583192&-0.23662&-0.0762139\\\hline
5s&-0.350374&-0.0969275&0.538291&-0.617767&-0.00442265&0.0923481&0.115864&-0.576655&-0.0108339\\\hline
6u&0.205348&0.302459&0.550447&0.0549231&-0.348898&0.420478&0.378317&0.56191&0.145699\\\hline
6s&0.105335&-0.155296&0.0193898&-0.283895&-0.0577008&0.220243&-0.31611&-0.296682&-0.0753326\\\hline
  \end{tabular}
\end{table*}

To learn the reward function from the observed trajectories based on the formulated MDP, we use the
coordinate and velocity direction of each grid as the feature, and learn the reward function
parameter from each set of data. The function approximator is a neural network with three hidden
layers and $[100,50,25]$ nodes.

We only test the proposed method with a neural-net function approximator, because it will take
prohibitive amount of time to learn the reward function with other methods, and the GP approach
relies on the set of supporting points. Assuming it takes only 100 iterations to converge, the
proposed method takes about one minute while others run for two to four weeks, and in practice, it
may take more iterations to converge.

To compare the reward function with and without stimulations, we adopt the same initial parameter
during reward function learning, and run both learning process with 10000 iterations with learning
rate 0.00001.

Given the learned reward function, we score the patient's recovery with the correlation coefficient
between the recovered rewards and the ideal rewards under the clinicians' instructions of the states
visited by the patient. The ideal reward for each state is the cosine similarity between the state's
velocity vector and the instructed direction.

The result is shown in Table \ref{tab:feature1}. It shows that the patient's ability to follow the
instructions is affected by the stimulations, but whether it is improved or not varies among
different directions. The clinical interpretations will be done by physicians.

\section{Conclusions}
\label{irl::conclusions}
This work deals with the problem of inverse reinforcement learning in large state spaces, and solves
the problem with a function approximation method that avoids solving reinforcement learning problems
during reward learning. The simulated experiment shows that the proposed method is more accurate and
scalable than existing methods, and can extends existing methods to high-dimensional spaces. A
clinical application of the proposed method is presented.

In future work, we will remove the requirement of a-priori known transition function by combining an
 environment model learning process into the function approximation framework.



\begin{thebibliography}{10}
\providecommand{\url}[1]{#1}
\csname url@samestyle\endcsname
\providecommand{\newblock}{\relax}
\providecommand{\bibinfo}[2]{#2}
\providecommand{\BIBentrySTDinterwordspacing}{\spaceskip=0pt\relax}
\providecommand{\BIBentryALTinterwordstretchfactor}{4}
\providecommand{\BIBentryALTinterwordspacing}{\spaceskip=\fontdimen2\font plus
\BIBentryALTinterwordstretchfactor\fontdimen3\font minus
  \fontdimen4\font\relax}
\providecommand{\BIBforeignlanguage}[2]{{%
\expandafter\ifx\csname l@#1\endcsname\relax
\typeout{** WARNING: IEEEtran.bst: No hyphenation pattern has been}%
\typeout{** loaded for the language `#1'. Using the pattern for}%
\typeout{** the default language instead.}%
\else
\language=\csname l@#1\endcsname
\fi
#2}}
\providecommand{\BIBdecl}{\relax}
\BIBdecl

\bibitem{irl::irl1}
A.~Y. Ng and S.~Russell, ``Algorithms for inverse reinforcement learning,'' in
  \emph{in Proc. 17th International Conf. on Machine Learning}, 2000.

\bibitem{irl::irl2}
P.~Abbeel and A.~Y. Ng, ``Apprenticeship learning via inverse reinforcement
  learning,'' in \emph{Proceedings of the twenty-first international conference
  on Machine learning}.\hskip 1em plus 0.5em minus 0.4em\relax ACM, 2004, p.~1.

\bibitem{irl::subgradient}
G.~Neu and C.~Szepesv{\'a}ri, ``Apprenticeship learning using inverse
  reinforcement learning and gradient methods,'' \emph{arXiv preprint
  arXiv:1206.5264}, 2012.

\bibitem{irl::maxentropy}
B.~D. Ziebart, A.~Maas, J.~A. Bagnell, and A.~K. Dey, ``Maximum entropy inverse
  reinforcement learning,'' in \emph{Proc. AAAI}, 2008, pp. 1433--1438.

\bibitem{irl::guidedirl}
C.~Finn, S.~Levine, and P.~Abbeel, ``Guided cost learning: Deep inverse optimal
  control via policy optimization,'' \emph{arXiv preprint arXiv:1603.00448},
  2016.

\bibitem{irl::relative}
A.~Boularias, J.~Kober, and J.~R. Peters, ``Relative entropy inverse
  reinforcement learning,'' in \emph{International Conference on Artificial
  Intelligence and Statistics}, 2011, pp. 182--189.

\bibitem{irl::kalman}
R.~Kalman and M.~M. C. B. D. R.~I. for Advanced Studies. Center~for
  Control~Theory, \emph{When is a Linear Control System Optimal?.}, ser. RIAS
  technical report.\hskip 1em plus 0.5em minus 0.4em\relax Martin Marietta
  Corporation, Research Institute for Advanced Studies, Center for Control
  Theory, 1963.

\bibitem{irl::maxmargin}
N.~D. Ratliff, J.~A. Bagnell, and M.~A. Zinkevich, ``Maximum margin planning,''
  in \emph{Proceedings of the 23rd international conference on Machine
  learning}.\hskip 1em plus 0.5em minus 0.4em\relax ACM, 2006, pp. 729--736.

\bibitem{irl::sequence}
Q.~P. Nguyen, B.~K.~H. Low, and P.~Jaillet, ``Inverse reinforcement learning
  with locally consistent reward functions,'' in \emph{Advances in Neural
  Information Processing Systems}, 2015, pp. 1747--1755.

\bibitem{irl::gaussianirl}
S.~Levine, Z.~Popovic, and V.~Koltun, ``Nonlinear inverse reinforcement
  learning with gaussian processes,'' in \emph{Advances in Neural Information
  Processing Systems 24}, J.~Shawe-Taylor, R.~S. Zemel, P.~L. Bartlett,
  F.~Pereira, and K.~Q. Weinberger, Eds.\hskip 1em plus 0.5em minus 0.4em\relax
  Curran Associates, Inc., 2011, pp. 19--27.

\bibitem{irl::pomdp}
J.~Choi and K.-E. Kim, ``Inverse reinforcement learning in partially observable
  environments,'' \emph{Journal of Machine Learning Research}, vol.~12, no.
  Mar, pp. 691--730, 2011.

\bibitem{irl::localirl}
S.~Levine and V.~Koltun, ``Continuous inverse optimal control with locally
  optimal examples,'' \emph{arXiv preprint arXiv:1206.4617}, 2012.

\bibitem{irl::deepirl}
M.~Wulfmeier, P.~Ondruska, and I.~Posner, ``Deep inverse reinforcement
  learning,'' \emph{arXiv preprint arXiv:1507.04888}, 2015.

\bibitem{irl::bayirl}
D.~Ramachandran and E.~Amir, ``Bayesian inverse reinforcement learning,'' in
  \emph{Proceedings of the 20th International Joint Conference on Artifical
  Intelligence}, ser. IJCAI'07.\hskip 1em plus 0.5em minus 0.4em\relax San
  Francisco, CA, USA: Morgan Kaufmann Publishers Inc., 2007, pp. 2586--2591.

\bibitem{irl::bayioc}
K.~Mombaur, A.~Truong, and J.-P. Laumond, ``From human to humanoid
  locomotion—an inverse optimal control approach,'' \emph{Autonomous robots},
  vol.~28, no.~3, pp. 369--383, 2010.

\bibitem{irl::multirl1}
C.~Dimitrakakis and C.~A. Rothkopf, ``Bayesian multitask inverse reinforcement
  learning,'' in \emph{European Workshop on Reinforcement Learning}.\hskip 1em
  plus 0.5em minus 0.4em\relax Springer, 2011, pp. 273--284.

\bibitem{irl::multirl2}
J.~Choi and K.-E. Kim, ``Nonparametric bayesian inverse reinforcement learning
  for multiple reward functions,'' in \emph{Advances in Neural Information
  Processing Systems}, 2012, pp. 305--313.

\bibitem{irl::rl}
R.~S. Sutton and A.~G. Barto, \emph{Reinforcement learning: An
  introduction}.\hskip 1em plus 0.5em minus 0.4em\relax MIT press Cambridge,
  1998, vol.~1, no.~1.

\bibitem{irl::BGI}
K.~{Li} and J.~W. {Burdick}, ``{Bellman Gradient Iteration for Inverse
  Reinforcement Learning},'' \emph{ArXiv e-prints}, Jul. 2017.

\bibitem{irl::motionvalue}
E.~Todorov, ``Linearly-solvable markov decision problems,'' in \emph{Advances
  in neural information processing systems}, 2007, pp. 1369--1376.

\bibitem{irl::sparsegaussian}
J.~Qui{\~n}onero-Candela and C.~E. Rasmussen, ``A unifying view of sparse
  approximate gaussian process regression,'' \emph{Journal of Machine Learning
  Research}, vol.~6, no. Dec, pp. 1939--1959, 2005.

\bibitem{irl::gaussianml}
C.~E. Rasmussen and C.~K. Williams, \emph{Gaussian processes for machine
  learning}.\hskip 1em plus 0.5em minus 0.4em\relax MIT press Cambridge, 2006,
  vol.~1.

\end{thebibliography}
\end{document}